# A case study on English-Malayalam Machine Translation

**Sreelekha. S**
IIT Bombay
India
sreelekha@cse.iitb.ac.in

**Pushpak Bhattacharyya**
IIT Bombay
India
pb@cse.iitb.ac.in

**Abstract**

In this paper we present our work on a case study on Statistical Machine Translation (SMT) and Rule based machine translation (RBMT) for translation from English to Malayalam and Malayalam to English. One of the motivations of our study is to make a three way performance comparison, such as, a) SMT and RBMT b) English to Malayalam SMT and Malayalam to English SMT c) English to Malayalam RBMT and Malayalam to English RBMT. We describe the development of English to Malayalam and Malayalam to English baseline phrase based SMT system and the evaluation of its performance compared against the RBMT system. Based on our study the observations are: a) SMT systems outperform RBMT systems, b) In the case of SMT, English - Malayalam systems perform better than that of Malayalam - English systems, c) In the case RBMT, Malayalam to English systems are performing better than English to Malayalam systems. Based on our evaluations and detailed error analysis, we describe the requirements of incorporating morphological processing into the SMT to improve the accuracy of translation.

## 1 Introduction

In a large multi-lingual society like India, there is a great demand for translation of documents from one language to another. Most of the state governments work is in the respective regional languages whereas the Union Government's official documents and reports are in bilingual form (English/Hindi). In order to have a proper communication there is a need to translate these documents and reports in the respective regional languages. The newspapers in regional languages are required to translate news in English received from International News Agencies. With the limitations of human translators most of this reports and documents are missing and not percolating down. A machine assisted translation system or a translator's workstation would increase the efficiency of the human translators. As is clear from above, India is rich in linguistic divergence there are many morphologically rich languages which are quite different from English as well as from each other, there is a great need for machine translation between them.

There are many ongoing attempts to develop MT systems for regional languages using various approaches (Kunchukuttan et al., 2014). The approaches to machine translation are categorized as, Rule Based or Knowledge Driven approaches and Corpus Based or Data-Driven approaches. The RBMT approaches are further classified into Transfer based MT, Interlingua MT and Dictionary based MT, while the Corpus Based approaches are classified into Example Based MT and SMT. In the case of English to Indian languages and Indian to Indian languages, there have been fruitful attempts with all approaches (Antony, 2013; Sreelekha et al., 2013; Sreelekha et al., 2014). This paper discusses various approaches used in English to Malayalam and Malayalam to English MT systems.

The rest of the paper is as follows, Section 2 deals with challenges in MT, Section 3 deals with approaches in MT, RBMT and SMT, Section 4 deals with Experiments conducted, Evaluations and Error analysis which concludes the main components of the paper.

## 2. Challenges in English–Malayalam MT

Major difficulties in Machine Translation are handling the structural difference between the two languages and handling the ambiguities.

### 2.1 Challenge of Ambiguity
There are three types of ambiguities: structural ambiguity, lexical ambiguity and semantic ambiguity.

### 2.1.1 Lexical Ambiguity

Words and phrases in one language often have multiple meaning in another language.

For example, the English sentence,

**English-** His view was good
**Malayalam-**
അവന്റെ അഭിപ്രായം നല്ലതായിരുന്നു
{ avante abhiprayam nallathayirunnu}

Here in the above sentence *"view"*, has ambiguity in meaning. It is not clear that whether the word *"view"*, is used as the *"opinion"* (*"അഭിപ്രായം"* {abhiprayam} in Malayalam) sense or the *"eye sight"* (*"കാഴ്ച"*{kazhcha} in Malayalam) sense. This kind of ambiguity has to be identified from the context.

### 2.1.2 Structural Ambiguity

In this case, due to the structural order, there will be multiple meanings. For example,

**Malayalam-**
അവിടെ വണ്ണമുള്ള പശുവും കാളയും ഉണ്ടായിരുന്നു
{avide vannamulla pashuvum kalayum undayirunnu}
**English- There were fat cows and buffalos there**

Here from the words *"വണ്ണമുള്ള പശുവും കാളയും"*{vannamulla pashuvum kalayum} it is clear that, cows are fat but it is not clear that buffallos are fat, since in Malayalam to represent fat cows and buffalos only one word *"വണ്ണമുള്ള"* {vannamulla} {fat} is being used. It can have two interpretations in English according to its structure.

*{There were fat cows and buffalos there}*
or
*{There were fat cows and fat buffalos there}*

To handle this kind of structural ambiguity is one of the big problems in Machine Translation.

### 2.1.3 Semantic Ambiguity

In this case, due to the understanding of the semantics, there will be multiple translations. For example, consider the English sentence,

*I eat with spoon and forks*
*I eat with my friends*

Here this English sentence can be translated in Malayalam as,

ഞാൻ സ്പൂണും ഫോർക്കും വച്ചാണ് കഴിക്കുന്നത്
*{njan spoonum forkum vachanu kazhikkunnathu}*
*{I spoons forks with eat }*
or
ഞാൻ എന്റെ സുഹൃത്തുക്കളുടെ കൂടെയാണ് കഴിക്കുന്നത്
*{njan ente suhruthukkalude koodeyanu kazhikkunnathu}*
*{ I my friends along with eat }*

Here, in the two English sentences *"with"* gets translated to വച്ചാണ് *{vachanu}* and കൂടെയാണ് *{koodeyanu}* respectively. This disambiguation requires knowledge to distinguish between *spoon- forks* and *friends*.

### 2.2 Structural Differences

There are word order differences between English and Malayalam such as, English language follows Subject -Verb- Object (SVO) and Malayalam language follows Subject-Object-Verb (SOV). The structural transfer between English- Malayalam is represented in figure 1.

Consider an example for word ordering,

**English-** Raman ate food
　　　　(S)　(V)　(O)
**Malayalam-** രാമൻ ഭക്ഷണം കഴിച്ചു
　　　　*{Raman bhakshanam kazhichu}*

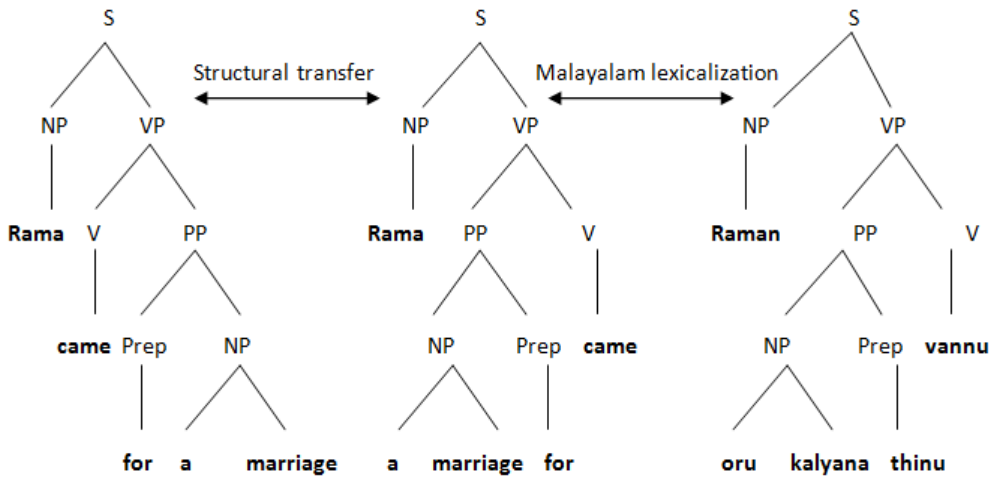

**Figure 1: Structural Transfer from English - Malayalam**

(S) (O) (V)

In addition, Malayalam is morphologically very rich as compared to English, wherein there are a lot of post-modifiers in the former as compared to the later.

For example, the word form *"കുട്ടിയുടെ"* *{kuttiyude} {of child}* is derived by attaching *"യുടെ"{yude}{of}* as a suffix to the noun *"കുട്ടി"{kutti}{child}* by undergoing an inflectional process. Malayalam exhibits agglutination of suffixes which is not present in English and therefore these suffixes has equivalents in the form of pre positions. For the above example, the English equivalent of the suffix *"യുടെ"* *{yude}* is the pre position *"of"* which is separated from the noun *"child"*.

This kind of structural differences have to be handled properly during translation.

### 2.3 Vocabulary Difference

Languages differ in the way they lexically divide the conceptual space and sometimes no direct equivalents can be found for a particular word or phrase of one language in another. Consider the sentence,

നാളെ കാവടിയാട്ടം ഉണ്ട്
*{ nale kavadiyattam undu}*
*{tomorrow kavadiyattam is there}*

Here the word, *"കാവടിയാട്ടം"* *{kavadiyattam}* as a verb has no equivalent in English, and this word have to be translated as *"the dance performed especially for the god Muruka using kavadi"*. Hence the sentence will be translated in English as,

*Tomorrow, the dance performed especially for the god Muruka using kavadi is there.*

Translating such language specific concepts pose additional challenges in machine translation.

### 3. Approaches of MT

One of the central design questions in machine translation is the syntactic structural transfer, which is the conversion from a syntactic analysis structure of the source language to the structure of the target language. The Vacquois triangle in the figure 2 depicts three different types of Machine Translation namely, Transfer based, Interlingua based and Statistical. They differ in the amount of linguistic processing performed before transferring concepts and structure from the source side to the target side. As can be seen Interlingua requires complete processing, Transfer based requires some and Statistical (a type of direct translation) requires none. The base of the triangle indicates the distance between the two languages and linguistic processing helps bridge the gap.

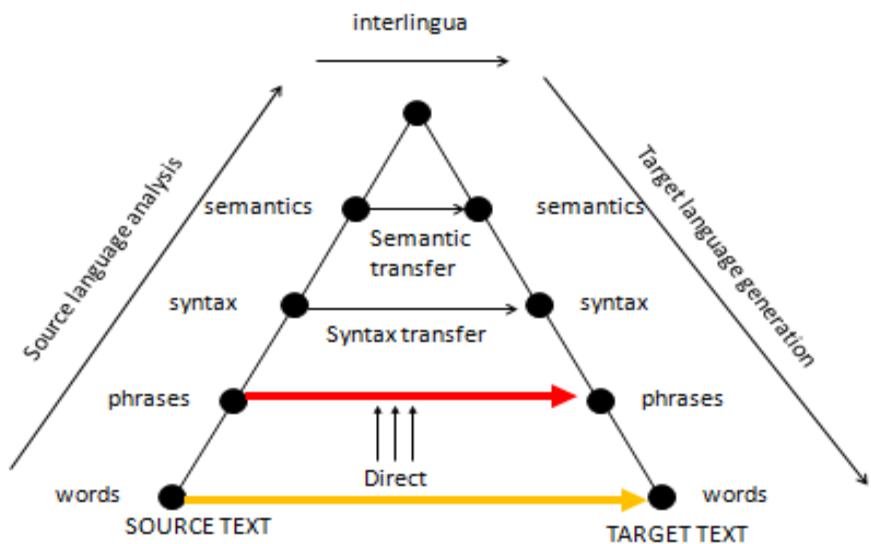

**Figure 2: Vacquois Triangle**

Direct translation is appropriate for structurally similar languages. Among the rule based approaches transfer based systems are more flexible and it can be easily extended to language pairs in a multilingual environment. The interlingua based systems can be used for multilingual translation since it used a language independent form. The Universal Networking Language has been proposed as the interlingua (Dave et al., 2002) for overcoming the language barrier.

## 3.1 Rule Based Machine Translation

### 3.1.1 Analysis

During this phase, from the input text information about the morphology, parts of speech, shallow phrases, entity and word sense disambiguation information is extracted.

### 3.1.2 Lexical transfer

The lexical transfer phase involves two parts namely word translation and grammar translation which is performed using high quality bilingual dictionary and transfer grammar rules.

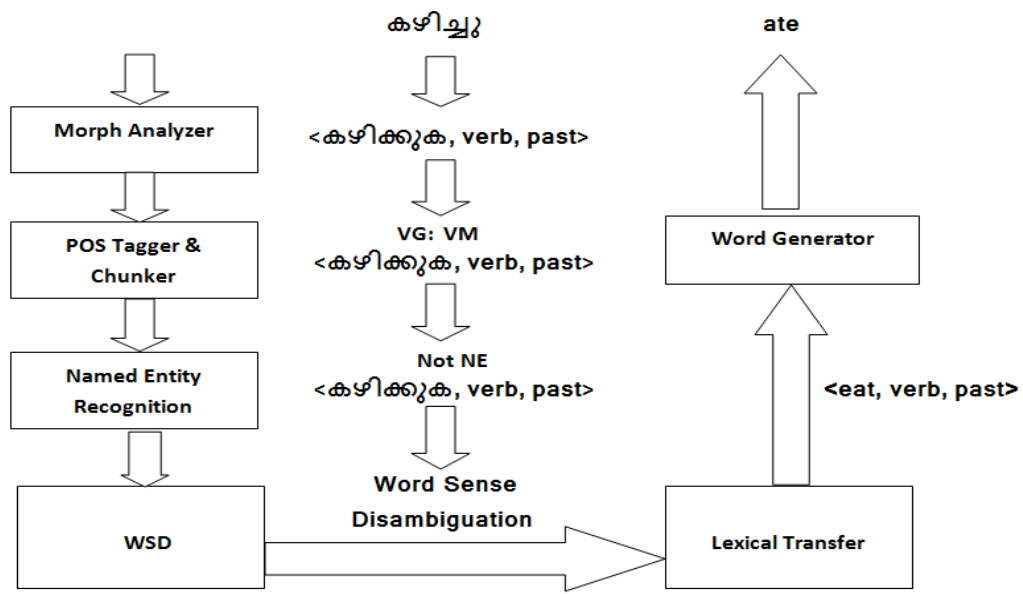

Figure 3: RBMT Functional flow

RBMT system (Sreelekha et..al. 2013) (Sunil et.al. 2011)(Latha et.al. 2012) requires a huge human effort to prepare the rules and linguistic resources, such as morphological analyzers, part-of-speech taggers and syntactic parsers, bilingual dictionaries, transfer rules, morphological generator and reordering rules etc. Specified rules for morphology play a major role in various stages of translation, such as syntactic processing, semantic interpretation and contextual processing of language. The transfer model involves three stages: analysis, transfer and generation. While translating a sentence RBMT system processes it word by word. The complete flow of translation of a word in the form of a pipeline is given in Figure 3.

### 3.1.3 Generation phase

Generation involves correction of the genders of the translated words since certain words are masculine in the source language but feminine in the target and vice versa. This is followed by short distance and long-distance agreements performed by intra-chunk and the inter-chunk modules concluded by word generation.

## 3.2 Statistical Machine Translation

Statistical models take the assumption that every word in the target language is a translation of the source language words with some probability (Brown et al., 1993). The words which have the highest probability will give the best translation. There are three different statistical approaches in MT, Word-based Translation, Phrase-based Translation, and Hierarchical phrase based model.

Consistent patterns of divergence between the languages (Dorr et al., 1994, Ramananthan et al., 2011, Kunchukuttan and Bhattacharyya 2012) when translating from one language to another, handling reordering divergence are one of the fundamental problems in MT.

We prepared a well aligned parallel corpus for training, testing and tuning as listed in the tables 1, 2, 3 and 4. As described in Figure 3 and section 2 English and Malayalam are structurally different there were difficulties during reordering. From experiments we

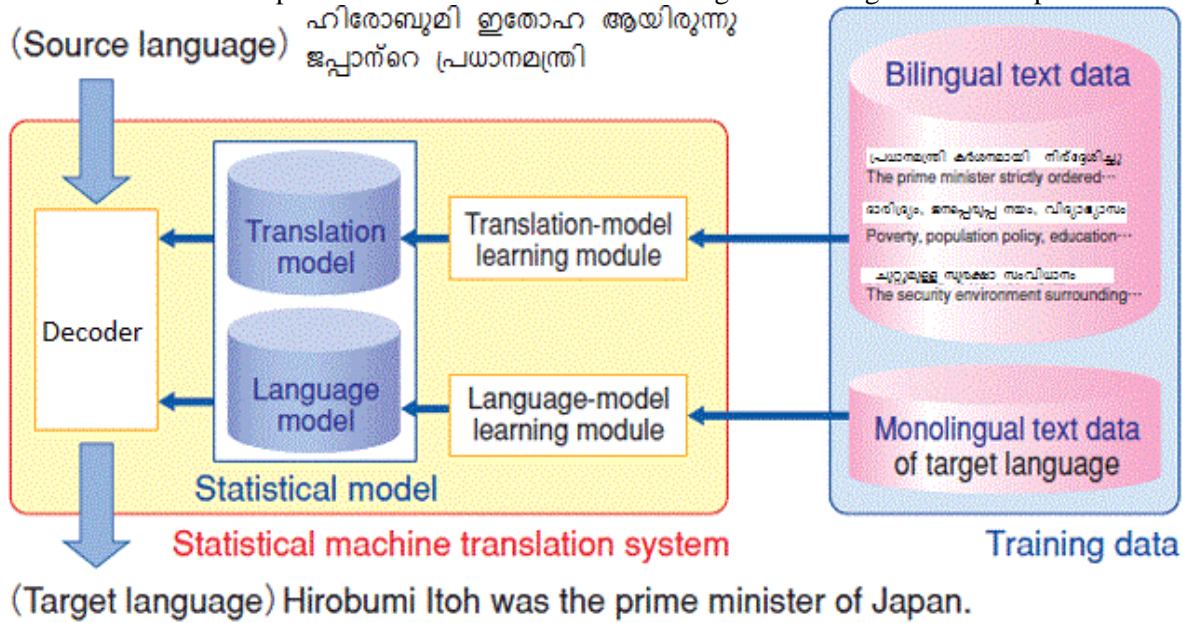

**Figure 4: SMT Functional flow**

Figure 4 shows the functional flow diagram of a SMT system. During training, from the parallel aligned sentences, word alignments and phrase alignments are learned. This leads to the extraction of phrases and thereby the phrase table, Translational model, Language Model, Distortion table etcetera is modeled. During decoding (Och and Ney, 2001; Och and Ney, 2003; Knight, 1999) the trained models will be decoded to generate the target language translations.

## 4 Experimental Discussion

### 4.1 SMT System Experiments

We now describe the development of our English- Malayalam and Malayalam-English SMT System[1]. The experiments performed and the comparisons with the results of the Rule Based system in the form of an error analysis is described in section 3.1. We use Moses (Koehn et al., 2007) and Giza++[2] for training and to generate the statistical models.

observed that SMT system fails to generate inflected word forms at many places since the system was unable to handle the rich morphology of Malayalam.

#### 4.1.1 English- Malayalam SMT

Consider an English sentence,
*He ate food with his friends.*
The English to Malayalam SMT system translated it as,
അവൻ അവന്റെ സുഹൃത്ത് ആഹാരം കഴിക്കുക
*{avan avante suhruth aaharam kazhikkuka }*
*{He his friend food ate}*
Even though the structural order was correct, here the SMT system is failed to generate the inflected form "*with his friends*" as "സുഹൃത്തുക്കളോടൊപ്പം"{*suhruthukkalodoppam*}, which is agglutinated with multiple suffixes, since these inflected word forms were absent in the training corpus. But the system translated "*ate*" as "കഴിക്കുക*{eat}* instead of the inflected past form "കഴിച്ചു"{*ate*}. Moreover enumerating all possibilities of inflected word forms is not possible manually. Hence the morphology limits the flexibility of SMT systems.

---
[1] http://www.cfilt.iitb.ac.in/SMT-EM/
[2] http://www.statmt.org/

### 4.1.2 Malayalam to English SMT

For example, consider a Malayalam sentence,
അവൻ അവന്റെ സുഹൃത്തുക്കളോടൊപ്പം ആഹാരം കഴിച്ചു
*{avan avante suhruthukkalodoppam aaharam kazhichu}*

*{He ate food along with his friends}*

The Malayalam to English SMT system translated it as,

He his *സുഹൃത്തുക്കളോടൊപ്പം food.* Here the system fails to translate the inflected form *"സുഹൃത്തുക്കളോടൊപ്പം" {suhruthukkalodoppam} {along with friends}.* Also the system missed to translate the word *"കഴിച്ചു" {kazhichu}* as "*ate*" since it couldn't find a matching inflected form in phrase table.

### 4.2 Rule-Based MT System Experiments

We have compared the SMT system translations with the RBMT system translations and the results are shown in the tables 5, 6, 7 and 8.

### 4.2.1 Malayalam - English RBMT

Consider the same Malayalam sentence,
അവൻ അവന്റെ സുഹൃത്തുക്കളോടൊപ്പം ആഹാരം കഴിച്ചു.
*{avan avante suhruthukkalodoppam aaharam kazhichu}*

The Malayalam to English RBMT system translated it as,

*He to along his friends ate food*

Here each of the words will be processed through the RBMT pipe line. As shown in figure 2, the important steps of the RB system flow for the word *"കഴിച്ചു"{kazhichu} {ate}* is,

**1. Analysis**: The morphological analyzer identifies the word *"കഴിച്ചു"{kazhichu}* as a verb in past tense. After POS tagging, it is identified that the word is a Main Verb and the Chunker determines that it is a part of a Verb Group. After WSD the appropriate sense is determined.

**2. Transfer:** The lexical transfer module translates it to "eat".

**3. Generation:** Since the sentence is short the

| Sl.No | Corpus Source | Training Corpus [Manually cleaned and aligned] | Corpus Size [Sentences] |
|---|---|---|---|
| 1 | ILCI | Tourism | 23750 |
| 2 | ILCI | Health | 23750 |
| 3 | Joshua | Tourism | 29518 |
| | Total | | 77018 |

Table 1: Statistics of Training Corpus

| Sl. No | Corpus Source | Tuning corpus [Manually cleaned and aligned] | Corpus Size [Sentences] |
|---|---|---|---|
| 1 | ILCI | Tourism | 250 |
| 2 | ILCI | Health | 250 |
| | Total | | 500 |

Table 2: Statistics of Tuning Corpus

| Sl. No | Corpus Source | Testing corpus [Manually cleaned and aligned] | Corpus Size [Sentences] |
|---|---|---|---|
| 1 | ILCI | Tourism | 1000 |
| 2 | ILCI | Health | 1000 |
| | Total | | 2000 |

Table 3: Statistics of Testing Corpus for BLEU score

| Sl. No | Corpus Source | Testing corpus [Manually cleaned and aligned] | Corpus Size [Sentences] |
|---|---|---|---|
| 1 | ILCI | Tourism | 50 |
| 2 | ILCI | Health | 50 |
| | Total | | 100 |

Table 4: Statistics of Testing Corpus for subjective evaluation

agreement phenomenon is not so significant. The word generator takes the information about "past tense" to give the final word form: "*ate*".

However the translation is far from good, considering that the translation of സുഹൃത്തുക്കളോടൊപ്പം {*suhruthukkalodoppam*} {along with friends} is "*to along his friends,*" which is not accurate. Here the system is not able to accurately determine the correct translation sense of "ഓടൊപ്പം"{*odoppom*}{*along with*}by splitting it into "ഓട്" {*odu*} {*to*} , "ഒപ്പം" {*oppam*}{*along*} leading to a poor lexical choice instead of "*along with*".

Consider the English sentence,
*He ate food along with his friends.*
The RBMT output for this sentence is,

അവൻ കഴിച്ചു ആഹാരം കൂട്ടുകാരുടെ
*{avan kazhihcu aaharam koottukarude}*
*{He ate food friend's}*

Here in the transfer stage *"along with"* is wrongly translated into "കാരുടെ"*{karude}* and it completely destroys the meaning of the sentence.

We observed that, although rule based MT was able to handle rich morphology, leading to meaning transfer, it was unable to effectively handle the appropriate translation and generation of function words and common word senses which are handled well by SMT, which improve fluency (Ahsan, et al. , 2010). As can be seen from the above described example, the translation of a single word requires a number of steps, each involving considerable linguistic inputs. Hence, RBMT process is extremely time consuming, difficult, and fails to analyze accurately and quickly a large corpus of unrestricted text due to inherent errors in the modules which are part of the system.

### 4.3 Evaluation

We have used subjective evaluation to determine fluency (F), an indicator of correct grammatical constructions present in the translated sentence and adequacy (A), an indicator of the amount of meaning being carried over from the source to the target. We did consider BLEU scores (Papineni et al.) also for evaluation. For each translation we assigned scores between 1 and 5 depending on how much sense the translation made and its grammatical correctness. The basis of scoring is given below:

- 5: If the translations are perfect.
- 4: If there are one or two incorrect translations and mistakes.
- 3: If the translations are of average quality, barely making sense.
- 2: If the sentence is barely translated.

| English- Malayalam MT System | Adequacy | Fluency |
|---|---|---|
| Rule Based | 55.6% | 47% |
| Statistical | 77.23% | 87% |

Table 5: Results of English- Malayalam SMT Vs. RBMT Subjective Evaluation

| English-Malayalam MT System | BLEU Score |
|---|---|
| Rule Based | 20.8 |
| Statistical | 39.90 |

Table 6 : Results of English- Malayalam SMT Vs. RBMT BLEU score

| Malayalam- English MT System | Adequacy | Fluency |
|---|---|---|
| Rule Based | 64.6% | 51% |
| Statistical | 74.89% | 85.34% |

Table 7: Results of Malayalam- English SMT Vs. RBMT Subjective Evaluation

| Malayalam- English MT System | BLEU Score |
|---|---|
| Rule Based | 29.9 |
| Statistical | 37.90 |

Table 8 : Results of Malayalam- English SMT Vs. RBMT BLEU score

- 1: If the sentence is not translated or the translation is gibberish.

Let S1, S2, S3, S4 and S5 be the counts of the number of sentences with scores from 1 to 5 and N be the total number of sentences evaluated. The formula (Bhosale et al., 2011) used for computing the scores is:

$$A/F = 100 * \frac{(S5 + 0.8 * S4 + 0.6 * S3)}{N}$$

We consider only the sentences with scores above 3. Moreover we penalize the sentences with scores 4 and 3 by multiplying their count by 0.8 and 0.6 respectively so that the estimate of scores is much better. As these scores are subjective, they vary from person to person in which case an inter annotator agreement is required. Since we had only one evaluator we do not give these scores. The results of our evaluations are given in Table 4 and Table 6.

### 4.4 Error Analysis

We have evaluated the translated outputs of both RBMT and SMT systems. The detailed error analysis for sentences exhibiting a variety of linguistic phenomena is shown in Tables 9 and 10. The result of BLEU score evaluation is displayed in Tables 5, 7 and the result of Subjective evaluation is displayed in Tables 6, 8.

It is clear from the evaluations that SMT outperforms RBMT. The reason that the SMT system had a very high fluency was due to plentiful evidences of good quality phrase pairs recorded in the phrase table. Moreover the language model used, helped in generating more natural translations. Also SMT which cannot split suffixes by itself was unable to handle the translation of suffix words in some cases. RBMT being able to use the morph analyzer, can easily separate the suffixes from the inflected words and generate translations inflected with correct gender number person, tense, aspect and mood (GNPTAM). However due to poor quality Word Sense Disambiguation incorrect translations are generated. This is mitigated by SMT since it records phrase translations with respect to frequency which acts as a more natural sense disambiguation mechanism.

Also we have observed that, the score of English-Malayalam translation quality is higher than that of Malayalam English translation. Malayalam is morphologically richer than English and Malayalam have more agglutinative suffixes attached as explained in Section 2.2, while in English it is not present. Therefore these Malayalam suffixes have English equivalents in the form of pre positions. English word can align to the words with agglutination in Malayalam easily, since it is a single word. But on the other hand while aligning form Malayalam -English the agglutinative word can map to a single word only, there is a chance to miss out the pre position or either the root word mapping, as it is separate words. Hence the translation quality of English - Malayalam SMT will be high as compared to Malayalam - English SMT.

Moreover, Malayalam to English RBMT performs better than English to Malayalam

| Sr. No. | | Malayalam- English MT - Sentence | Explanation of phenomena |
|---|---|---|---|
| 1 | Source ML Sentence | രോഗി വളരെയധികം ക്ഷീണിക്കുന്നതു കൊണ്ട് ഭാരം കൂടുതലായി തോന്നുന്നു. | The SMT translation have fluency and adequacy. SMT translation only one mistake of insertion case "even". The RBMT failed in conveying meaning and to follow grammatical structure. Because of the word order and structure the meaning changed. |
| | Meaning | Patient becomes very weak and thinks that she puts upon weight | |
| | RBMT | Patient is very weak she puts upon weight and thinks a lot. | |
| | SMT | On being very weak patient thinks her weight to be too much even. | |
| 2 | Source Malayalam Sentence | കൊതുക് കാരണം ഉണ്ടാകുന്ന ഈ രോഗം യാതൊരു കുത്തിവയ്പ്പോ ഉചിതമായ ചികിത്സയോ ഇതുവരെ കണ്ടുപിടിച്ചിട്ടില്ല. | The RBMT system doesn't follow grammatical structure and meaning is lost. Also it faces deletion problem. Didn't translate all words. Also ambiguity in translating the word "കാരണം" and translated as "reason" in place of "because" The SMT output is good except some prepositions and conjunctions, one missing word. IT can be framed in better way. |
| | Meaning | Any vaccine or proper cure of this disease casued by mosquitoes is not known till now . | |
| | RBMT system | Mosquitos reason the disease vaccination treatment not found | |
| | Statistical System | The disease because of virus not even any vaccine or proper treatment not found | |

Table 9 : Malayalam-English SMT, RBMT- Error Analysis

| | Sr. No. | English-Malayalam MT - Sentence | Explanation of phenomena |
|---|---|---|---|
| 1 | Source English Sentence | Tus located on the banks of the Berach river near Udaipur and the Sun temple have an important place in the study of sculpting tradition. | The SMT translation is very good both in terms of fluency and adequacy. The RBMT translation have problems in meaning translation, to follow grammatical structure, ordering, vocabulary problem and ambiguity problem while translating "river bank" and it translated as "ബാങ്കു"{bank}. In SMT the conjunction form "and" is missing and also the inflected suffixes. The RBMT handles the inflections to a level. SMT have problems in handling prepositional phrases and inflections of content words. |
| | Rule based system | ടുസ് ബെരക് നദിയുടെ ബാങ്കുകളിലെ അടുത്ത ഉദയ്പൂർ സ്ഥാനം കണ്ടുപിടിച്ചു , സൂര്യന്റെ ക്ഷേത്രത്തില് സുല്പ്റ്റിംഗ് പാര-ന്പര്യത്തിന്റെ അധ്യനത്തില് ഒരു പ്രധാനപ്പെട്ട സ്ഥാനം ഉണ്ട്. | |
| | Meaning | Tus bank of berach river near Udaipur found place, in sun temple in sculpting traditional study one important place | |
| | Statistical System | ഉദയ്പൂര് അടുത്ത ബേടച്ച് നദി സ്ഥിതി ചെയ്യുന്ന റ്റൂസ് സൂര്യക്ഷേത്രത്തിനും ശില്പ കല പഠനം പ്രധാനപ്പെട്ട സ്ഥാനം. | |
| | Meaning | Tus located bedach river near Udaipur sun temple and sculpting study important place | |
| 2 | Source English Sentence | In 1886 the national central museum was established during the visit of the Prince of Wales and in 1986 was opened for the public. | The RBMT translation have grammatical structure problem. Vocabulary problem, the insertion case "അറിയപ്പെട്ട", 'museum' has been translated wrongly as "ദേശീയം". Even though handles inflections conjuctions are missing in RBMT. SMT output conveys the meaning and structure. Only problems in translating conjunction "and". Also to translate the verb phrase "opened for" correctly with suffix. |
| | Rule based system | 1886 ഇല് കേന്ദ്ര കാഴ്ചബംഗ്ലാവ് 1986 ഇലും വള്സിലും രാജകുമാരന്റെ സന്ദര്ശനത്തിന്റെ സമയത്ത് അറിയപ്പെട്ട ആ ദേശീയം പൊതുജനത്തിന് വേണ്ടി തുറക്കപ്പെട്ടു. | |
| | Meaning | In 1886 central museum in 1986 in wales recognized during prices visit that national opened for public | |
| | Statistical System | 1886-ൽ വെയിൽസ് രാജകുമാരന്റെ സന്ദര്ശനവേളയിൽ നാഷണൽ മ്യൂസിയം സ്ഥാപിക്കപ്പെട്ടത് 1986-ൽ ജനങ്ങൾ തുറന്നു കൊടുത്തു. | |
| | Meaning | In 1886 during wales prince visit national museum established in 1986 public opened | |

Table 10 : English- Malayalam SMT, RBMT- Error Analysis

RBMT. Since Malayalam-English require Morphology analysis and English to Malayalam RBMT requires Morphology Generation. During Malayalam Morphology Analysis, from a single inflected word, agglutinated suffixes are getting separated and it is easy to identify equivalent group words and to translate during lexical transfer. But on the other hand during Morphology generation while generating a single inflected Malayalam word from a group of English words, all words may not get properly formed. There is higher chance to get error in generation of equivalent Malayalam inflected form. Thus Malayalam to English RBMT can handle inflections more accurately than English to Malayalam RBMT.

## 5 Conclusion

In this paper we have mainly focused on the comparative performance of Statistical Machine Translation and Rule- Based Machine Translation. Our major observations are,
1. Translation quality of SMT is relatively high as compared to the RBMT system, considering that the efforts required to build RBMT systems is huge.
2. SMT perform better for English to Malayalam systems with a bleu score of 39.90 with a fluency of 87 % and adequacy of 77.23% comparing to Malayalam to English systems with a bleu score of 37.90, fluency of 85.34% and adequacy of 74.89%.
3. RBMT performs better for Malayalam to English with a bleu score of 29.9 and an adequacy of 64.6%, fluency of 51%, as of English to Malayalam with a bleu score of 20.8 and adequacy of 55.6%, fluency of 47%.

As discussed in the experimental section, SMT, although lacks the ability to handle rich morphology, does not fall much behind RBMT. It has a staggering advantage over RBMT in terms of fluency and the ability to capture natural structure (Sreelekha et.al. 2013). This leads to the requirement of incorporating morphological processing into SMT for generating quality Machine Translations.

Our future work will be focused on the integration of Morphological processing into the Statistical Machine Translation system and thereby develop a better MT system.

## Acknowledgments

This work is funded by Department of Science and Technology, Govt. of India under Women Scientist Scheme- WOS-A with the project code- SR/WOS-A/ET-1075/2014.

## References

Ananthakrishnan Ramananthan, Pushpak Bhattacharyya, Karthik Visweswariah, Kushal Ladha, and Ankur Gandhe. 2011. *Clause-Based Reordering Constraints to Improve Statistical Machine Translation.*IJCNLP, 2011.

Anoop Kunchukuttan and Pushpak Bhattacharyya. 2012. *Partially modelling word reordering as a sequence labeling problem, COLING 2012*.

Anoop Kunchukuttan Abhijit Mishra, Rajen Chatterjee, Ritesh Shah and Pushpak Bhattacharyya, Shata-Anuvadak: Tackling Multiway Translation of Indian Languages, LREC 2014, Rekjyavik, Iceland.

Antony P. J. 2013. *Machine Translation Approaches and Survey for Indian Languages,* The Association for Computational Linguistics and Chinese Language Processing, Vol. 18, No. 1, March 2013, pp. 47-78

Arafat Ahsan, Prasanth Kolachina, Sudheer Kolachina, Dipti Misra Sharma and Rajeev Sangal. 2010. *Coupling Statistical Machine Translation with Rule-based Transfer and Generation.* amta2010.amtaweb.org

Bonnie J. Dorr. 1994. *Machine Translation Divergences: A Formal Description and Proposed Solution.*Computational Linguistics, 1994.

Franz Josef Och and Hermann Ney. *A Systematic Comparison of Various Statistical Alignment Models.* Computational Linguistics, 2003.

Franz Josef Och and Hermann Ney. 2001. *Statistical Multi Source Translation.* MT Summit 2001.

Ganesh Bhosale, Subodh Kembhavi, Archana Amberkar, Supriya Mhatre, Lata Popale and Pushpak Bhattacharyya. 2011. *Processing of Participle (Krudanta) in Marathi.* ICON 2011, Chennai, December, 2011.

Kevin Knight. 1999. *Decoding complexity in word-replacement translation models*, Computational Linguistics, 1999.

Kishore Papineni, Salim Roukos, Todd Ward and Wei-Jing Zhu. 2002. *BLEU: a Method for Automatic Evaluation of Machine Translation*, Proceedings of the 40th Annual Meeting of the Association for Computational Linguistics (ACL), Philadelphia, July 2002, pp. 311-318.

Latha R. Nair, David Peter S, Renjith Ravindran. 2012. *Design and Development of a Malayalam to English Translator- A Transfer based Approach*, International Journal of Computational Linguistics, Volume(3): Issue(1), 2012.

Peter E Brown, Stephen A. Della Pietra. Vincent J. Della Pietra, and Robert L. Mercer*. 1993. The Mathematics of Statistical Machine Translation: Parameter Estimationn. *ACL 1993*.

Philipp Koehn, Hieu Hoang, Alexandra Birch, Chris Callison-Burch, Marcello Federico, Nicola Bertoldi, Brooke Cowan, Wade Shen, Christine Moran, Richard Zens, Chris Dyer, Ondrej Bojar, Alexandra Constantin, Evan Herbst. 2007. *Moses: Open Source Toolkit for Statistical Machine Translation*, Annual Meeting of the ACL, demonstration session, Prague, Czech Republic, June 2007.

Sreelekha, Pushpak Bhattacharyya, Malathi D 2014. *Lexical Resources for Hindi – Marathi MT*, WIDRE Proceedings, LREC 2014..

Sreelekha, Raj Dabre, Pushpak Bhattacharyya 2013. *Comparison of SMT and RBMT, The Requirement of Hybridization for Marathi – Hindi MT* ICON, 10[th] International conference on NLP, December 2013.

Shachi Dave, Jignashu Parikh and Pushpak Bhattacharyya. 2002. *Interlingua based English-Hindi Machine Translation and Language Divergence* , JMT 2002.

Sunil R, Nimtha Manohar, Jayan V, KG Sulochana. 2011, *Development of Malayalam Text Generator for translation from English*, India Conference (INDICON), 2011 Annual IEEE